\documentclass[conference]{IEEEtran}
\usepackage{cite}
\usepackage{amsmath,amssymb,amsfonts}
\usepackage{algorithmic}
\usepackage{graphicx}
\usepackage{textcomp}
\usepackage{xcolor}
\def\BibTeX{{\rm B\kern-.05em{\sc i\kern-.025em b}\kern-.08em
    T\kern-.1667em\lower.7ex\hbox{E}\kern-.125emX}}

\makeatletter
\def\ps@IEEEtitlepagestyle{%
  \def\@oddfoot{\mycopyrightnotice}%
  \def\@evenfoot{}%
}
\def\mycopyrightnotice{%
  {\footnotesize 979-8-3195-3697-6/26/\$31.00~\copyright~2026 IEEE\hfill}
  \gdef\mycopyrightnotice{}%
}
\newcommand{\linebreakand}{%
  \end{@IEEEauthorhalign}
  \hfill\mbox{}\par
  \mbox{}\hfill\begin{@IEEEauthorhalign}
}
\makeatother

\begin{document}

\title{MoTop: Motion-Topological Model For Micro AU Detection}

\author{
\IEEEauthorblockN{Huai-Qian Khor}
\IEEEauthorblockA{\textit{CMVS, University of Oulu} \\
Oulu, Finland \\
huai.khor@oulu.fi}
\and
\IEEEauthorblockN{Mengting Wei}
\IEEEauthorblockA{\textit{CMVS, University of Oulu} \\
Oulu, Finland \\
Mengting.Wei@oulu.fi}
\and
\IEEEauthorblockN{Yante Li}
\IEEEauthorblockA{\textit{AII, CAAS}, Beijing, China \\ \textit{Zhongyuan Research Center, CAAS}, Henan, China\\
liyante@caas.cn}
\linebreakand
\IEEEauthorblockN{Chu Kiong Loo}
\IEEEauthorblockA{\textit{University of Malaya} \\
Kuala Lumpur, Malaysia \\
ckloo.um@um.edu.my}
\and
\IEEEauthorblockN{Guoying Zhao\thanks{Corresponding author: guoying.zhao@oulu.fi}\IEEEauthorrefmark{1}}
\IEEEauthorblockA{ \textit{ELLIS Institute Finland}, Espoo, Finland\\
\textit{CMVS, University of Oulu}, Oulu, Finland \\
guoying.zhao@oulu.fi}
}
\maketitle

\def\thefootnote{}
\footnotetext{$^{*}$Corresponding author.}

\begin{abstract}
Facial micro-expressions are spontaneous, brief, and subtle facial movements that reveal suppressed emotions in high-stakes environments. In contrast to classic expression analysis, detecting action unit (AU) yields a finer representation of facial movements, serving as a preliminary step before defining expression classes and other downstream tasks. Therefore, it represents a crucial upstream task in facial analysis, and improving an AU detection module increases the precision of facial analysis. Despite that, detecting AU is challenging because of the constrictive nature of the AU activation regions, leading to confusion among different AUs known as AU ambiguity. To model the fine-scale changes, we propose \textbf{MoTop}, a motion-topological model that is augmented with a learnable motion context, yielding regional soft guidance for facial activity, followed by facial landmarks that capture the fine-scale topological changes of micro AUs. To increase the micro facial landmark representations, we amplify the encoded facial landmark transitions via linear extrapolation, thereby increasing the spatial proximity of landmarks and enhancing the low-intensity landmark dynamics. In addition, we design anatomical facial clusters that enhance the hierarchical representation, facilitating multi-scale modelling of facial geometry and improving micro-topological representations. With these contributions, we have achieved state-of-the-art performance on the CD6ME protocol for the micro AU detection task.

\end{abstract}


\vspace{-0.5em}
\section{Introduction}
\label{sec:intro}

Facial micro-expression is a spontaneous, subtle and rapid facial motion that is elicited in high-stakes environments \cite{ekman1971constants} \cite{ekman2009lie}, such as psychological diagnosis and interviews. In computational analysis, Action Unit (AU) detection serves as a fine-grained, first-level analysis that discretises facial muscle movements and is ultimately utilised for downstream categorisation and insight mining. For micro-expression AUs, they serve as the fundamental task in identifying elicited emotions \cite{yan2014casme} \cite{davison2016samm}, whereby trained annotators first identify the occurrence of each micro AU with the naked eye, followed by classifying the elicited emotion based on the collectively occurring micro AUs. Computationally, detecting 12 multi-label micro AUs approximates the equivalent labour-intensive task, in which the micro AUs are selected based on their sample size \cite{varanka2023data} and their significance in downstream tasks such as emotion recognition \cite{li2022deep}.

However, detecting micro AU is highly challenging. Firstly, AU ambiguity \cite{khor2025infused} is a prominent problem in which two different AUs occur close together, causing the model to fail to distinguish between them. Secondly, the minuscule facial motions of micro-expressions exhibit extremely subtle AU activity regions; with limited changes in spatial proximity, the model fails to learn the facial dynamics, especially at the node level. These two problems are tightly coupled and mutually reinforced, exacerbating the difficulty of micro-AU detection.

\begin{figure}[!t]
    \centering
    \includegraphics[width=0.8\columnwidth]{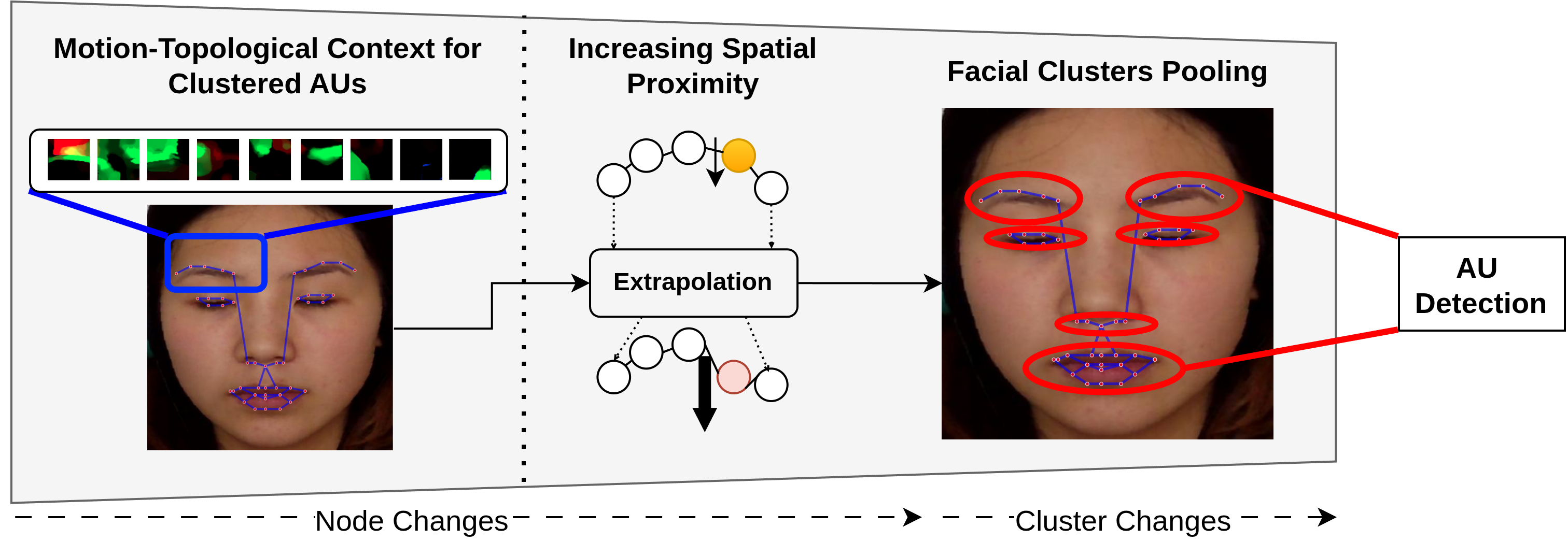}
    \vspace{-1em}
    \caption{A conceptual diagram depicting the integration of motion context into MoTop to alleviate AU ambiguity, where the motion context provides a regional representation while the landmarks yield a finer one. A linear extrapolation is computed to amplify the spatial proximity of facial nodes. The anatomical facial clusters then encapsulate the topological changes. The eventual multi-scale facial dynamic then contributes to the micro AU detection task.}
    \vspace{-1.5em}
    \label{fig:conceptualdiagram}
\end{figure}

In micro AU detection, there were several attempts \cite{khor2025infused} \cite{zhang2025regulatory} \cite{varanka2023data} \cite{varanka2023learnable} on learning the micro AU activity; despite that, AU ambiguity (AU 1 (inner brow raiser) and AU 4 (brow lowerer) may have concurrent and overlapping appearances because they are both located around the eyebrows' region) remain unsolved which potentially causes feature learning ambiguity. Meanwhile, the Graph Convolutional Network (GCN) has achieved considerable success in expression recognition tasks \cite{li2022deep}\cite{liu2022graph}\cite{wei2023geometric}\cite{zhang2025micro} by modelling inter-landmark relationships. However, direct adaptation to micro AU detection is challenging due to the low-intensity dynamics of micro AUs and the low spatial proximity among micro facial landmarks, resulting in minuscule facial landmark transitions. Coupled with AU ambiguity, it exacerbates the difficulty of learning landmark dynamics.

To leverage the representativeness of both optical flow and facial topological transitions, we propose \textbf{MoTop}, a motion-topological model that jointly learns the optical flow and landmark transitions for AU detection. Given the minuscule micro facial landmark dynamics, we first amplify the encoded facial node transitions via linear extrapolation, thus increasing the spatial proximity of facial landmarks and enhancing low-intensity dynamics. We then inject the motion context, derived from optical flow, into MoTop, providing regional guidance to steer facial landmarks for fine-scale micro-AU detection. Additionally, to obtain a higher-level graph representation, we design anatomical facial clusters around each facial organ to build a hierarchical representation (from node-level to organ-level) for learning node transitions, enabling multi-scale modelling of facial geometry. Figure \ref{fig:conceptualdiagram} demonstrates our approach to resolving the gradual sensitivity to changes in fine-scale micro AUs.


In summary, these are our contributions for the micro AU detection task.
\begin{itemize}
    \item Amplifying the encoded facial landmark transitions via a linear extrapolation strategy, thereby increasing the proximity of facial nodes and enhancing the representation of low-intensity micro AU dynamics.

    \item Injecting the encoded motion context into graph node representations, yielding regional guidance towards fine-scale and complex micro AU activation. 

    \item Designing anatomical facial clusters that enhance hierarchical representation and facilitate multi-scale modelling of facial geometry.
\end{itemize}
\section{Methodology}
Our proposed framework, MoTop, utilises both optical flow images and 2D facial landmarks to learn motion-topological changes. It is first injected with an encoded motion context, fine-tuned from the Motion Context Network (MCN), yielding regional guidance for facial landmarks to encode the fine-scale regional changes. The facial landmarks are amplified via a linear extrapolation technique, thereby increasing the proximity of facial nodes and enhancing the dynamics of low-intensity micro AUs. Then, a graph pooling based on the anatomical facial clusters encapsulates the motion-topological transition, creating a multi-scale facial geometry that contributes to fine-scale AU detection. To complement the missing representation of MoTop, a separate motion-only AU Detection Network (ADN) is fine-tuned to fuse with the multi-scale motion-topological features from MoTop.

\subsection{Preliminaries}
\begin{figure*}[!ht]
    \centering
    \includegraphics[width=0.7\linewidth]{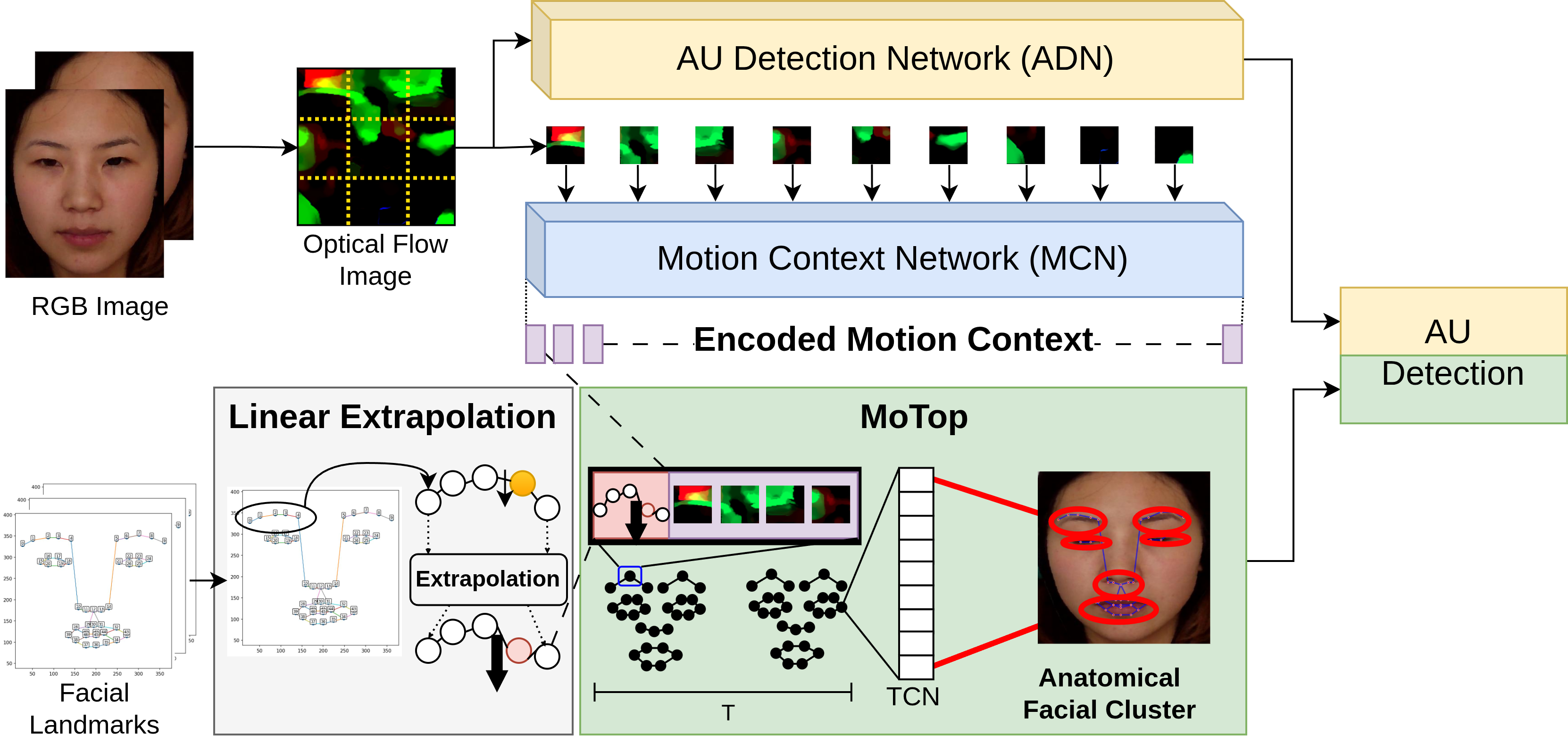}
    \caption{An illustration of our proposed framework, MoTop. It demonstrates the process of injecting fine-tuned motion-context embeddings from the Motion Context Network (MCN) into the MoTop node embeddings. The linear extrapolation also amplifies landmark node transitions, increasing spatial proximity and enhancing the node transitions. The eventual AU Detection fuses results from both the AU Detection Network (ADN), which is directly fine-tuned with the optical flow image, and MoTop, which yields motion-topological features that model fine-scale micro AU transitions.}
    \vspace{-1.5em}
    \label{fig:framework}
\end{figure*}

\noindent\textbf{Optical Flow Representation}. The input for both motion networks, the AU Detection Network (ADN) and the Motion Context Network (MCN), is the optical flow image. As demonstrated in Equation \ref{eq:of}, the optical flow changes \cite{sun2010secrets} and its derivative, known as optical strain \cite{shreve2009towards}, $\epsilon$ demonstrated in Equation \ref{eq:strain_short}, are computed between the onset and apex frames. 

\vspace{-0.5em}
\begin{equation}
	u = [u, v]^T ,\qquad  \epsilon = \frac{1}{2}[\nabla u + (\nabla u)^T]
    \label{eq:strain_short}
\end{equation}

\vspace{-0.5em}

\begin{equation}
    I_{OF} = (d_x, d_y, \epsilon)
    \label{eq:of}
\end{equation}

\noindent\textbf{Topological Representation}. For facial topological changes, we extract 68 facial landmarks \cite{bulat2017far}, then filter out landmarks that are unlikely to reflect facial motion, such as the facial boundary and nose bridge landmarks, resulting in 47 landmarks. The learnable adjacency matrix of nodes is designed to model intra-node changes within each facial part \cite{wei2023geometric}.

\noindent\textbf{GCN-TCN}. The architecture of our proposed framework, MoTop, combines a Graph Convolutional Network (GCN) and a Temporal Convolutional Network (TCN). Based on Equation \ref{eq:gcn}, GCN encapsulates node-to-node connections among facial landmarks via a learnable adjacency matrix (edges), followed by a TCN to extract temporal changes in nodes between successive frames, as shown in Equation \ref{eq:tcn}. The temporal learner is accompanied by a skip connection to alleviate vanishing-gradient problems, given that changes in the nodes can be extremely subtle for micro AUs. In our framework, we utilise four successive GCN-TCN blocks with dimensions 32, 64, 64, and 128.

\vspace{-1em}
\begin{equation}
    x^{'}_{i} = \Theta^{T} \sum^{J}_{j\in N(i) \cup \{i\}} \frac{e_{j, i}}{\sqrt{\hat{d}_j\hat{d}_i}} x_j
    \label{eq:gcn}
\end{equation}

\noindent where $\Theta^{T}$ is the convolutional filter matrix, $J$ denotes the number of edges, $e_{j, i}$ denotes the edge weight from $j$ to $i$, $\frac{1}{\sqrt{\hat{d}_j\hat{d}_i}}$ denotes the normalization term for edges, and $x_j$ is the input node.
\begin{equation}
    x^{t}_{i} = \sigma(TCN(\sigma(x^{'}_{i, t})) + TCN(x_{i, t}))
    \label{eq:tcn}
\end{equation}
\noindent where $\sigma$ denotes the ReLU activation.

\subsection{MoTop}
\noindent\textbf{Linear Extrapolation}. Given the low-intensity dynamics of micro AUs and the proximity of facial nodes, landmark changes are insignificant for feature extraction; therefore, encoding and amplifying node changes are required to extract a meaningful topological representation. To encode the node changes, we calculate the difference between two successive frames as $\mathrm{diff}(f_{n+1}, f_{n})$, where $n \in \mathbb{N}$ and $\mathbb{N}$ is the number of frames. Inspired by LatentMag \cite{wei2025latentmag}, we amplify the representation by extrapolating the difference between two consecutive frames, as shown in Equation \ref{eq:extrapolation}. Formally, $f_{ext}$ represents the extrapolated frame between two neighbouring frames; in our context, we obtain the encoded extrapolation between onset-apex and apex-offset frames. $\alpha$ denotes the extrapolation factor that magnifies changes in the landmark frame. 
\begin{equation}
    f_{ext} = f_{n} + \alpha * \mathrm{diff}(f_{n+1}, f_{n}), 
\qquad n \in \mathbb{N}, \; \mathbb{N} = 3
    \label{eq:extrapolation}
\end{equation}

\noindent\textbf{Motion Tokens}. In MoTop, motion tokens provide a soft regional context that complements landmarks' inability to distinguish genuine facial motions from signal noise. Theoretically, each motion token is associated with a landmark based on its approximate spatial coordinates, enabling it to capture the localised motion context relevant to that landmark. Therefore, we propose the Motion Context Network (MCN) with a TinyViT \cite{wu2022tinyvit} backbone, where the motion context, $M_{47}$, is created via a 1D adaptive average pooling, tallying to the number of facial landmark nodes used in MoTop. $M_{47}$ is passed to a fully connected layer, and focal loss is utilised to optimise $M_{47}$ towards the AU ground truth. This enables encoding an AU-responsive motion context for our MoTop framework.

With the motion context, $M_{47}$, it is then attached to the normalised facial landmark coordinates (Equation \ref{eq:extrapolation}) as input features to MoTop, as demonstrated in Equation \ref{eq:feature_input}, where the input dimension in total is 130 ( 2 Landmarks + 128 motion context embedding ). As such, MoTop can detect regional variations and localised, fine-scale landmark changes.

\vspace{-1em}
\begin{equation}
    I_{top} = \{f_{ext, x}, f_{ext, y}, M_{47}\}
    \label{eq:feature_input}
\end{equation}

\vspace{-0.5em}
\noindent\textbf{Anatomical Facial Clusters}. To model a hierarchical feature that summarises motion-topological fine-scale changes, we design facial clusters using anatomical heuristics to perform graph average pooling on the last GCN-TCN layer. In our heuristic, we define a total of six clusters, which include: (1) right brow, (2) left brow, (3) right eye, (4) left eye, (5) nose tip, (6) mouth, facial regions where AUs are likely to happen. We perform graph-based global mean pooling over the GCN-TCN outputs, summarising 47 nodes into six facial clusters represented by $r^t_{no, c}$, where $no$ denotes nodes, $c$ clusters, and $t$ temporality. We then compute the cluster temporal changes and predict the AUs by averaging over the temporal dimension $t$ followed by a FC layer. MoTop is then fine-tuned with another focal loss with AU groundtruth labels.



\vspace{-0.5em}
\subsection{Micro AU Detection}

\noindent\textbf{Patch-based AU Detector}. To complement the potential lack of representation in MoTop, where landmarks are missing from some AU-active regions such as the cheek, we propose training a separate motion-only AU Detection Network (ADN) to fuse with MoTop. Similarly, we adopt TinyViT \cite{wu2022tinyvit} as the ADN backbone due to the conscriptive nature of micro AU and the patch embeddings learnt from the transformer-based network are comparatively more sensitive than those of a convolutional network. The optimisation is performed using a separate focal loss that trains directly on $I_{OF}$, and its output is then used in the AU detection task via decision-level fusion.

\noindent\textbf{Decision-level Fusion}. With reference to Figure \ref{fig:framework}, ADN drives patch-embedding-based learning, while MoTop extracts fine, multi-scale features. We then perform decision-level fusion for the final AU detection task, as demonstrated in Equation \ref{eq:classification}, to leverage the strengths of both networks.

\vspace{-1.0em}

\begin{equation}
    \vspace{-0.5em}
    \hat{y} = \frac{1}{T}\sum_{t=1}^{T}{r^t_{no,c}} + ADN(I_{OF})
    \label{eq:classification}
\end{equation}

\noindent Given that ADN, MCN, and MoTop are tasked with learning separate types of features, we fine-tune each model with its own binary cross-entropy focal loss. As such, ADN learns to optimise optical flow features to detect AU; MCN optimises the 47-motion context, $M_{47}$, for each AU; whereas MoTop optimises the $I_{top}$ to model regional, fine-scale variations that are AU-sensitive. 
 
\begin{equation}
    \vspace{-0.5em}
    \mathcal{L} = Focal(\hat{AU}, AU) + Focal(M_{47}, AU) + Focal(r^t_{no,c}, AU)
    \vspace{-0.5em}
\end{equation}

\section{Experiments}
\subsection{Dataset \& Protocol}
The datasets that are utilised in our experiments are CASME \cite{yan2013casme}, CASME II \cite{yan2014casme}, CAS(ME)$^{3}$ \cite{li2022cas}, 4DME \cite{li2022deep}, MMEW \cite{ben2021video} and SAMM \cite{davison2016samm}. As proposed in \cite{varanka2023data}, we combine these six databases with 12 overlapping AUs, since the definition of emotional labels can vary across databases. We utilised the protocol known as Composite Database 6 Micro-Expressions (CD6ME) \cite{varanka2023data}, which performs a leave-one-database-out cross-validation. In detail, one database will be opted out as a test set while the rest will be used for training. The process is repeated for each database until all databases have been tested once, after which their Macro-F1 Scores are averaged.

\subsection{Experimental Setup}
In our proposal, three separate models were optimised. The experiments were carried out with an Adam optimiser with a learning rate of 0.001 for all models. For learning rate decay, both ADN and MCN used an exponential learning rate decay (ExponentialLR) with a gamma of 0.9. For MoTop, a multistep learning rate decay (MultiStepLR) was used with a gamma of 0.9, with decay every 20 epochs. The experiments were implemented with PyTorch and the PyTorch Geometric library.

\subsection{Comparison with State-of-the-art}
To validate our findings, we benchmark our method against other works evaluated under the CD6ME protocol, using the average Macro-F1 Score.  
\begin{table*}[!ht]
    \centering
    \caption{Benchmarking table compares the performance of the proposed algorithm with baseline results, with the percentage of samples across AUs listed. \textbf{Bolded} entries represent the best while \underline{underlined} entries are the second best performing results. }    
    \vspace{-1em}
    \footnotesize
    \scalebox{0.8}{
    \begin{tabular}{|c|c|c|c|c|c|c|c|c|c|c|c|c|c|c|}
        \hline
        Method  & AU1 & AU2 & AU4 & AU5 & AU6 & AU7 & AU9 & AU10 & AU12 & AU14 & AU15 & AU17 & Average\\
        & (12\%) & (11\%) & (28\%) & (5\%) & (2\%) & (10\%) & (5\%) & (3\%) & (7\%) & (11\%) & (2\%) & (3\%) & \\   
        \hline
        \hline
        LBP-TOP \cite{varanka2023data} & 0.4160 & 0.3670 & 0.6200 & 0.0000 & 0.0000 & 0.0000 & 0.0170 & 0.0000 & 0.0000 & 0.0350 & 0.0000 & 0.0000 & 0.1210 \\   

        Off-ApexNet \cite{gan2019off} & \textbf{0.7490} & 0.7020 & \underline{0.8630} & 0.1350 & 0.0330 & 0.4450 & \textbf{0.3660} & 0.1830 & 0.3200 & 0.3770 & 0.1900 & 0.3820 & 0.3950 \\
        
        SSSNet \cite{varanka2023data} & \underline{0.7460} & 0.7210 & \textbf{0.8750} & 0.1360 & 0.0530 & 0.4860 & 0.2030 & 0.1930 & 0.3620 & \underline{0.4080} & 0.2270 & \underline{0.4470} & 0.4050 \\     

        ResNet10 \cite{varanka2023data} & 0.6870 & 0.6520 & 0.8380 & 0.0830 & 0.0640 & 0.3960 & 0.1100 & 0.0680 & 0.3100 & 0.2960 & 0.0880 & 0.3960 & 0.3320 \\
        
        ResNet18 \cite{varanka2023data} & 0.6018 & 0.6541 & 0.6595 & 0.1612 & \underline{0.1645} & 0.4362 & 0.2227 & 0.2378 & 0.3828 & 0.3298 & 0.3152 & 0.4276 & 0.3828 \\

        ResNet34 \cite{varanka2023data} & 0.7210 & 0.7170 & 0.8580 & 0.1120 & 0.0540 & 0.3870 & 0.1600 & 0.1350 & 0.3650 & 0.3990 & 0.1480 & 0.3920 & 0.3710 \\

        ResNet50 & 0.6474 & 0.6841 & 0.7525 & 0.1015 & 0.0228 & 0.3825 & 0.1561 & 0.0476 & 0.2970 & 0.2476 & 0.2160 & 0.2877 & 0.3202 \\

        LED \cite{varanka2023learnable} & 0.5270 & 0.4570 & 0.6370 & 0.0790 & 0.0070 & 0.1930 & 0.1360 & 0.0850 & 0.2650 & 0.3670 & \underline{0.3300} & 0.3170 & 0.2830 \\

        ViT-Tiny \cite{wu2022tinyvit} & 0.6924 & \underline{0.7353} & 0.7984 & 0.2129 & 0.1510 & \underline{0.4785} & 0.2371 & 0.1987 & 0.3742 & 0.3877 & 0.2871 & 0.3999 & 0.4128 \\
          
        InfuseNet \cite{khor2025infused} &
        0.6997 & 0.6804 & 0.7921 & 0.1929 & 0.1572 & 0.4438 & \underline{0.3263} & \textbf{0.2987} & \underline{0.3987} & 0.3523 & 0.2973 & 0.4364 & \underline{0.4230} \\
        RFT \cite{zhang2025regulatory} & 0.5890
         & 0.5710  & 0.7720 & \textbf{0.4480} & 0.0068 & 0.3420 & 0.3090 & \underline{0.2500} & 0.3450 & \textbf{0.4750} & 0.3670 & 0.3250 & 0.4050 \\
        
		\hline
        

         \textbf{MoTop} & 0.7426
         & \textbf{0.8022} & 0.8583 & \underline{0.2566} & \textbf{0.1896} & \textbf{0.4992} & 0.2436 & 0.2484 & \textbf{0.4318} & 0.3743 & \textbf{0.4348} & \textbf{0.4507} & \textbf{0.4610} \\

        \hline 
    \end{tabular}

    }  
    \label{tab:benchmarking_table}
\end{table*}
Based on Table \ref{tab:benchmarking_table}, MoTop achieves the best performance in 6 of 12 AU classes while maintaining comparable performance in the others. In detail, compared with InfuseNet and RFT, we observe improvements in the AU 2 (Outer Brow Raiser), AU 12 (Lip Corner Puller), and AU 17 (Chin Raiser) regions, which exhibit significant AU ambiguity. This demonstrates the topological model's ability to learn localised motion features, in contrast to relying solely on regional motion descriptors such as optical flow, by leveraging motion context. Meanwhile, as facial landmarks lack information on the cheek region, the topological learner struggles with AUs that occur near the cheek region, such as AU 9 (Nose Wrinkler), AU 10 (Upper Lip Raiser) and AU 14 (Dimpler). In conjunction with that, MoTop utilises not only motion context but also a separate motion classifier to perform decision-level fusion for AU detection. This enables fine-grained motion changes and the inclusion of the cheek regions in the AU detection task; notably, AU 6 (Cheek Raiser) shows a slight improvement over other cheek-related AUs. In general, MoTop alleviates some facial ambiguity while retaining comparable performance across most AUs, except for AU 9, due to persistent, over-challenging ambiguity.

\subsection{Ablation Study}

To study the efficacy of motion context and facial clusters, we perform an ablation study on MoTop as shown in Table \ref{tab:ablation_motop}. Based on Table \ref{tab:ablation_motop}, we determine that landmarks without motion context are insufficient for topological learning. The addition of facial clusters can increase performance from 0.4151 to 0.4284 (\textcolor{green}{+1\%}) by introducing another level of feature hierarchy via graph mean pooling. In addition to ADN, it yields improvement from 0.4284 to 0.4610 (\textcolor{green}{+3\%}) as the combination of modules provides a more well-rounded set of features for the eventual AU detection task. However, for ADN alone, performance drops from 0.4610 to 0.4075 (\textcolor{red}{-5\%}) when motion-topological features are excluded. This demonstrates that the MoTop can extract fine-scale motion-topological transitions, and that ADN complements the model by providing potentially missing representations.

\begin{table}[!ht]
    \centering
    \vspace{-2em}
    \caption{Ablation study of MoTop to dissect the separate efficacy of motion context, facial clusters and motion detector.}
    \vspace{-1em}
    \resizebox{0.7\columnwidth}{!}{
        \begin{tabular}{|c|c|c|c|c|}
        \hline
             Landmarks & Motion Context & Facial Clusters & Motion Detector & MF1 \\
             \hline
             \checkmark & X & X & X & 0.0812 \\
             \checkmark & \checkmark & X & X & 0.4151\\
             \checkmark & \checkmark & \checkmark & X & 0.4284\\
             \checkmark & \checkmark & \checkmark & \checkmark & \textbf{0.4610}  \\
         \hline
        \end{tabular}
    }
    \label{tab:ablation_motop}
    \vspace{-2em}
\end{table}

\begin{figure}
    \centering
    \includegraphics[width=0.65\linewidth]{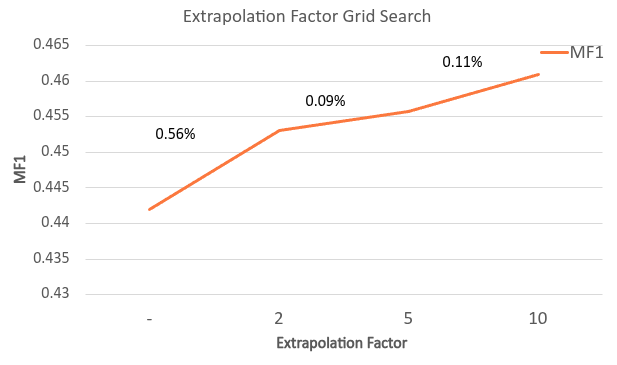}
    \caption{MF1 variation with an expanding extrapolation factor, with the gradient(\%) of performance growth visualised. The growth of MF1 with increasing factor decreases from 0.56\% to 0.11\%, which indicates that the optimum factor lies below 10.}
    \label{fig:extrapolation_factor}
    \vspace{-1em}
\end{figure}

Additionally, we study the sensitivity of the extrapolation factor across the ranges of 2, 5, and 10 using MoTop. As shown in Figure \ref{fig:extrapolation_factor}, a higher extrapolation factor may yield more significant performance gains despite minimal changes. The reason is that a higher extrapolation factor increases the magnitude of node changes; as a result, the topological learner is more sensitive to learning larger node displacements for AU detection.

\subsection{Qualitative Analysis}

\begin{figure}[!ht]
    \centering
    \vspace{-1em}
    \includegraphics[width=0.8\linewidth]{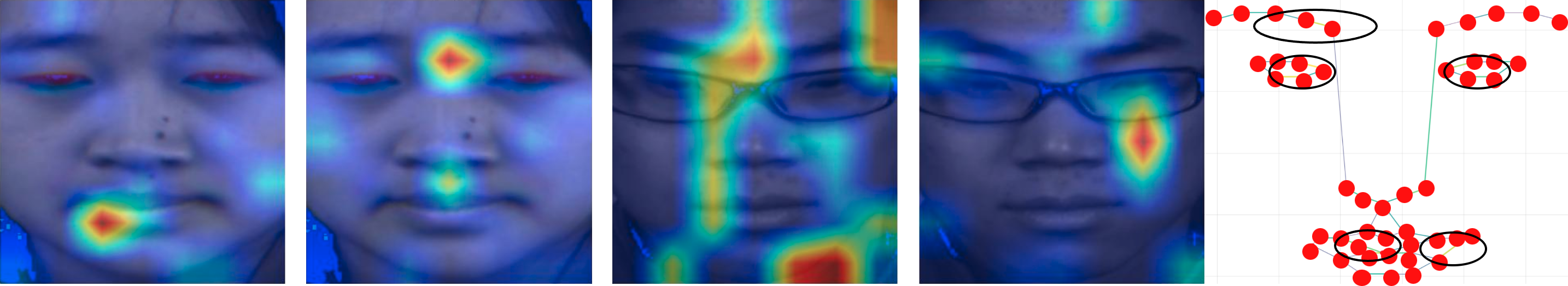}
    \caption{Illustration of Grad-CAM for the correct detection of AU14 (Dimpler), AU15 (Lip Corner Depressor), AU4 (Brow Lowerer), and AU7 (Lid Tightener). A visualised GCN adjacency matrix of the last GCN-TCN block.}
    \vspace{-0.5em}
    \label{fig:gradcam}
\end{figure}

\noindent\textbf{Activation Maps}. To understand the synergy between the motion and the motion-topological detector, we utilise Grad-CAM \cite{selvaraju2017grad} to visualise the activation maps of the TinyVIT (motion detector) and also visualise the adjacency matrix of the last GCN block. Based on Figure \ref{fig:gradcam}, the patch-based methodology of the motion detector can extract localised motion patches, as observed in the first, second, and fourth Grad-CAM images. The third Grad-CAM image shows a potential case of AU ambiguity, but the inclusion of topological features alleviates it in the flow motions, leading to the correct detection of AU 4 (Brow Lowerer). In the same figure, we observe that the motion-topological part of MoTop encircles active regions around the eyebrow, eye, and certain mouth areas, thereby augmenting feature learning for crucial AU regions in parallel with MoTop's motion detection module.

\begin{figure}[!ht]
    \centering
    \vspace{-1em}
    \includegraphics[width=0.5\linewidth]{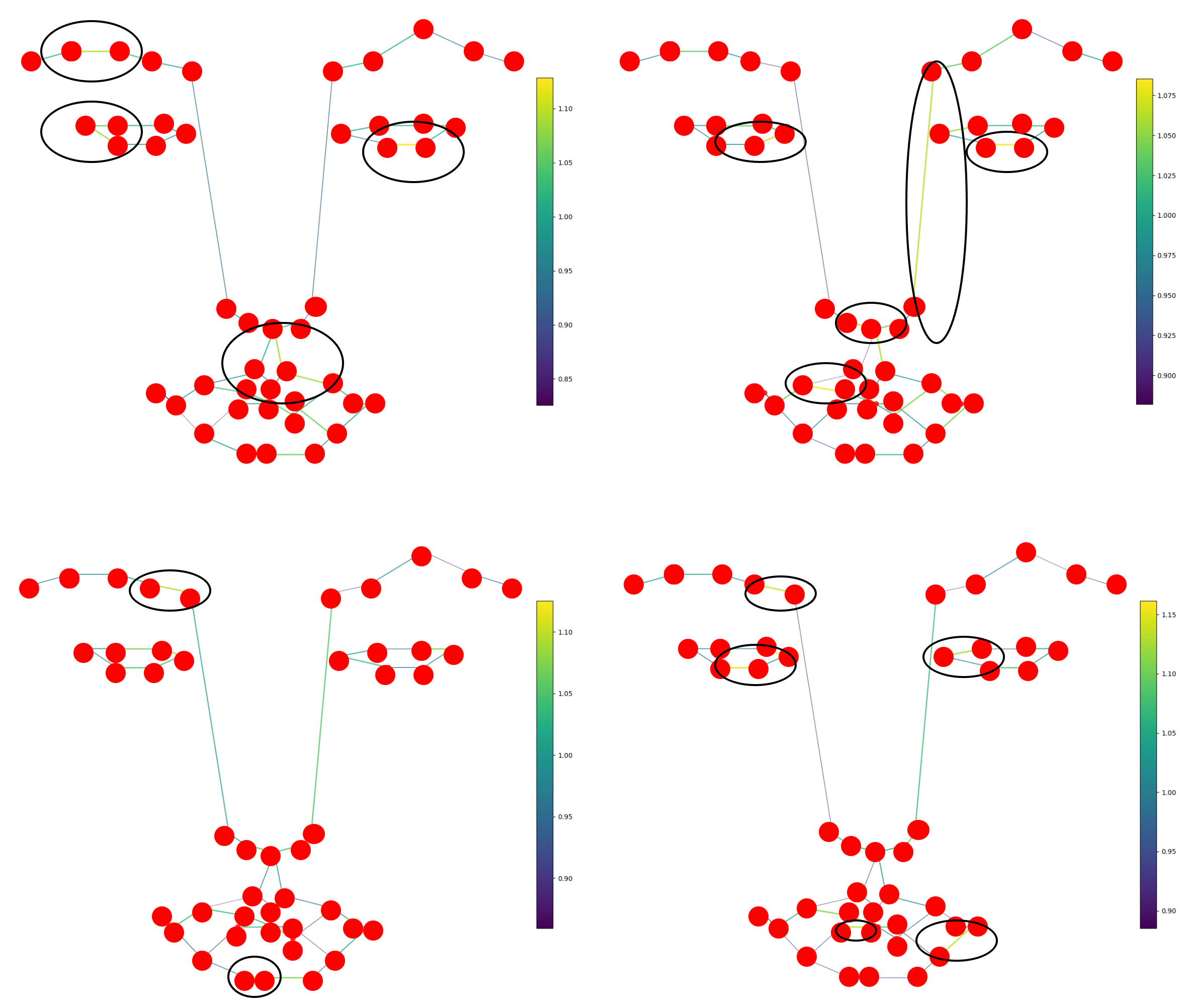}
    \caption{The illustration of adjacency matrix values or edges' weights for GCN 1, GCN 2, GCN 3 and GCN 4.}
    \vspace{-1em}
    \label{fig:gcn_transition}

\end{figure}

\noindent\textbf{Adjacency Weights Transition}. 
To study how the adjacency matrix transitions across four successive GCNs, we visualise the matrix values and highlight regions with higher weights, as shown in Figure \ref{fig:gcn_transition}. Based on the figure, we can observe that earlier GCNs emphasise regional and cross-region eye motion, for example, in GCN 2. The later layers, such as GCN 3 and GCN 4, focus on localised, balanced regions compared with earlier layers. Comparing both, we can deduce that the motion-topological features are gradually adapted to fine-scale changes as dimensionality increases, eventually yielding a motion-topological transition that is both localised and fine-scale, thereby alleviating the AU ambiguity problem.

\subsection{Detect-Then-Recognise (DTR) Task}
\begin{figure}
    \centering
    \includegraphics[width=0.5\linewidth]{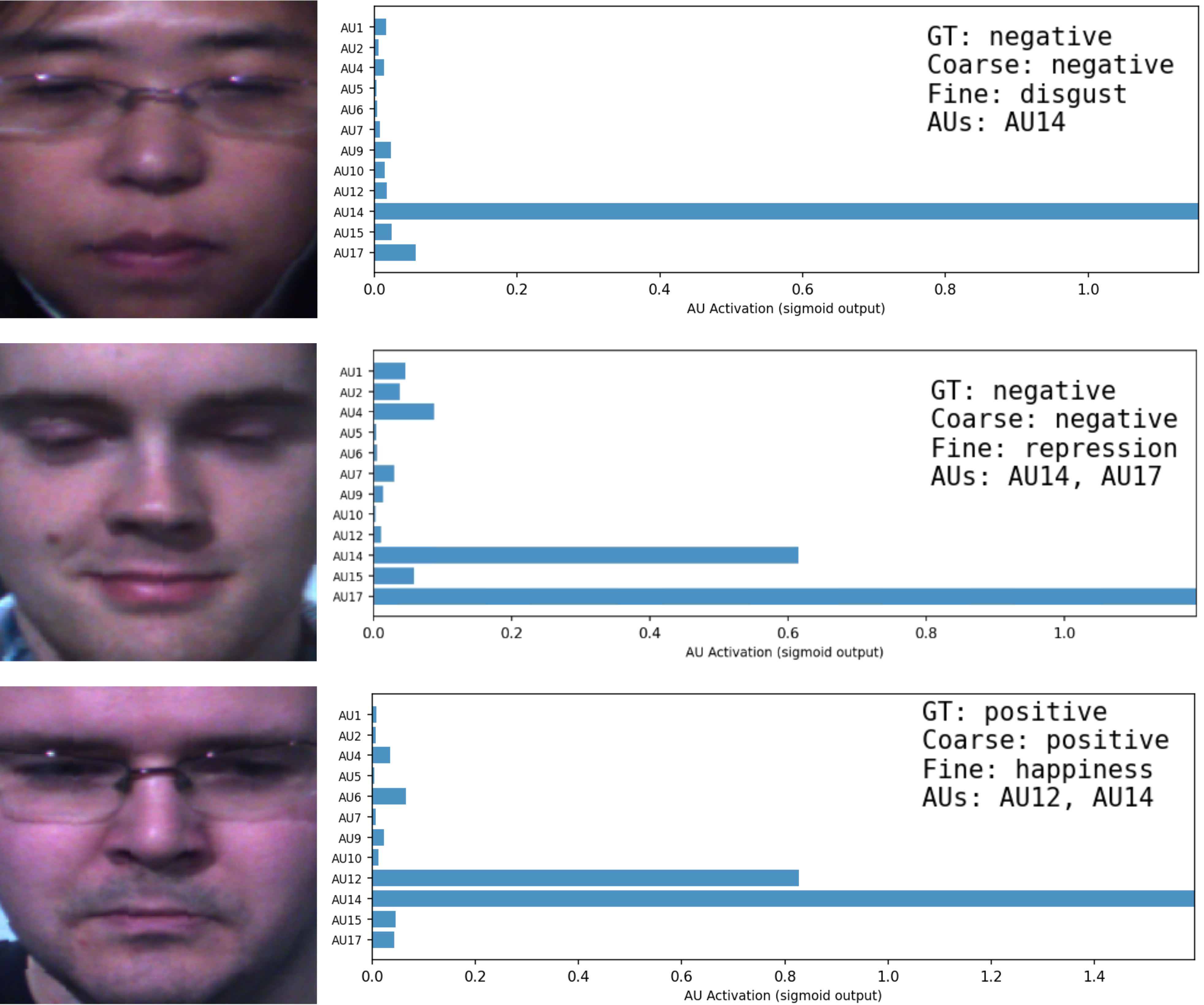}
    \caption{An example of the detect-then-recognise task (DTR) multi-granular emotion classes in SMIC dataset \cite{li2013spontaneous}.}
    \label{fig:DTR}
    \vspace{-2em}
\end{figure}

In human annotation, detecting micro facial action units (micro AUs) is crucial for identifying and interpreting micro-expressions. Motivated by this observation, we formulate a detect-then-recognise (DTR) paradigm that explicitly associates detected micro AUs with micro-expression categories, thereby mimicking the human annotation process while simultaneously evaluating the semantic quality of the detected AUs. Using the proposed MoTop framework on the SMIC dataset \cite{li2013spontaneous}, we estimate both coarse-grained emotions (three classes with available ground-truth labels) and fine-grained emotions (five classes without ground-truth annotations) from the detected micro AUs. The AU-to-emotion mappings follow the annotation protocols defined for SMIC \cite{li2013spontaneous} for coarse-grained categories and CASME II \cite{yan2014casme} for fine-grained categories. As illustrated in Figure \ref{fig:DTR}, the predicted emotions exhibit consistent semantic coherence across both granularities, indicating that MoTop can effectively infer micro-expressions from detected micro AUs on a completely unseen dataset. Notably, meaningful fine-grained emotion categories emerge despite the absence of fine-level annotations in the SMIC dataset. The precision in identifying fine-grained emotions demonstrates the efficacy of MoTop in fine-scale analysis and in alleviating potential AU ambiguity.

\section{Conclusion}
\vspace{-0.4em}

In our work, we have contributed to improving fine-scale micro AU detection with MoTop, a motion-topological AU detector that uses late fusion with the AU Detection Network (ADN). With respect to AU ambiguity, injecting motion context into MoTop's graph embeddings yields soft regional guidance, enabling facial landmarks to adapt to facial topological changes within that guidance, thereby modelling fine-scale motion-topological transitions. Given the presence of close-proximity nodes, we have computed a linear extrapolation to amplify landmark displacements and enhance the representation of low-intensity micro-AU dynamics. Additionally, we design anatomical facial clusters to encapsulate multiple node changes via graph mean pooling, facilitating multi-scale modelling of facial geometry and providing an improved representation of motion-topological transitions for studying micro AUs. By combining these multiple computations and modules, we improve upon the state of the art in the CD6ME micro AU detection protocol. Nevertheless, the task of micro AU detection remains challenging under the CD6ME protocol, where problems such as domain differences, imbalanced class learning, AU ambiguity, and minuscule AU activity persist.

\vspace{-0.5em}
\section*{Acknowledgements}
\vspace{-0.5em}
This work was supported by the Research Council of Finland Academy Prof. project EmotionAI (grants 336116, 359894), HPC project FaceCanvas (grant 364905), the Univ. of Oulu \& Research Council of Finland Profi 7 (grant 352788), and EU HORIZON-MSCA-SE-2022 project ACMod (grant 101130271). Y. Li was supported by CAAS-ASTIP-2026-AII, the Central Public-interest Fund (JBYW-AII-2026-30), and the Zhongyuan Research Center (ZYZX202503). We thank CSC - IT Center for Science, Finland, for computational resources.


\newpage

\bibliographystyle{IEEEtran}
\bibliography{strings}


\end{document}